\pdfoutput=1
\documentclass[11pt]{article}

\usepackage{emnlp2022}

\usepackage{times}
\usepackage{latexsym}

\usepackage[T1]{fontenc}
\usepackage[utf8]{inputenc}
\usepackage{graphicx}
\usepackage{booktabs}
\usepackage[inline]{enumitem}
\usepackage{makecell}

\usepackage{microtype}

\title{Cascading Biases: Investigating the Effect of Heuristic Annotation Strategies on Data and Models}
\author{Chaitanya Malaviya \and
  Sudeep Bhatia \and
  Mark Yatskar \\
  University of Pennsylvania \\
  \texttt{\{cmalaviy, bhatiasu, myatskar\}@upenn.edu} \\}

\begin{document}
\maketitle

\newcommand{\hname}{traces}

\begin{abstract}

Cognitive psychologists have documented that humans use cognitive heuristics, or mental shortcuts, %, in a range of problem-solving tasks \cite{simon1956rational, tversky1974judgment}.
%These heuristics allow us 
to make quick decisions while expending less effort.
While performing annotation work on crowdsourcing platforms,  %humans are similarly inclined to engage in heuristics \cite{eickhoff2018cognitive}.
we hypothesize that such heuristic use among annotators cascades on to data quality and model robustness.
% is a driver of data quality. % during crowdsourcing.
In this work, we study cognitive heuristic use in the context of annotating multiple-choice reading comprehension datasets. %, where annotators are given a passage, and asked to write a question with multiple options.
We propose tracking \textbf{\textit{annotator heuristic \hname}}, where we tangibly measure low-effort annotation strategies that could indicate usage of various cognitive heuristics.
We find evidence that annotators might be using multiple such heuristics, based on correlations with a battery of psychological tests. % measuring heuristic-seeking behavior. 
Importantly, heuristic use among annotators determines data quality along several dimensions: (1) known biased models, such as partial input models, more easily solve examples authored by annotators that rate highly on heuristic use, (2) models trained on annotators scoring highly on heuristic use don't generalize as well, and (3) heuristic-seeking annotators tend to create qualitatively less challenging examples. Our findings suggest that tracking heuristic usage among annotators can potentially help with collecting challenging datasets and diagnosing model biases.

\end{abstract}

\section{Introduction}

While crowdsourcing is an effective and widely-used data collection method in NLP, it comes with caveats. Crowdsourced datasets have been found to contain artifacts from the annotation process, and models trained on such data, can be brittle and fail to generalize to distribution shifts \cite{gururangan2018annotation, kaushik-lipton-2018-much, mccoy2019right}. In this work, we ask whether systematic patterns in annotator behavior influence the quality of collected data. 

We hypothesize that usage of \textit{cognitive heuristics}, which are mental shortcuts that humans employ in everyday life, can cascade on to data quality and model robustness. For example, an annotator asked to write a question based on a passage might not read the entire passage or might use just one sentence to frame a question. 
% Previous work sugggests that annotators can be susceptible to systematic use of such heuristics \cite{eickhoff2018cognitive} in relevance judgement tasks, which causes poorer label accuracy. 
Annotators may seek shortcuts to economize on the amount of time and effort they put into a task. This behavior in annotators, characterized by examples that are acceptable but not high-quality, can be problematic.%, especially in large-scale data collection.

We analyze the extent to which annotators engage in various low-effort strategies, akin to cognitive heuristics, by tracking indicative features from their annotation data in the form of \textbf{annotator heuristic \hname}. First, we crowdsource reading comprehension questions where we instruct workers to write hard questions. Inspired by research on human cognition \cite{simon1956rational, tversky1974judgment}, we identify several heuristics that could be employed by annotators for our task, such as satisficing \cite{simon1956rational}, availability \cite{tversky1973availability} and representativeness \cite{kahneman1972subjective}. We measure their potential usage by featurizing the collected data and annotation metadata (e.g., time spent and keystrokes entered) (\S\ref{sec:heuristics}). Further, we identify instantiations of these heuristics that correlate well with psychological tests measuring heuristic thinking tendencies in humans, such as the cognitive reflection test ~\cite{frederick2005cognitive, toplak2014assessing, sirota2021measuring}. 
Our psychologically plausible measures of heuristic use during annotation can be aggregated per annotator, forming a holistic summary of the data they produce.
% These psychologically plausible profiles provide a basis for grouping examples by possible neural processing taking place in annotators brains. 

Based on these statistics, we analyze differences between examples created by annotators who engage in different levels of heuristics use. Our first finding is that examples created by strongly heuristic-seeking annotators are also easier for models to solve using heuristics (\S\ref{sec:bias}). We evaluate models that exploit a few known biases and find that examples from annotators who use cognitive heuristics are more easily solvable by biased models. We also examine what impact heuristics have on trained models. Previous work \cite{geva-etal-2019-modeling} shows that models generalize poorly when datasets are split randomly by annotators, likely due to the existence of artifacts. We replicate this result and find that models generalize even worse when trained on examples from heuristic-seeking annotators.

To understand which parts of the annotation pipeline contribute to heuristic-seeking behavior in annotators, we also tease apart the effect of components inherent to the task (e.g., passage difficulty) as opposed to the annotators themselves (e.g., annotator fatigue) (\S\ref{sec:influencers}).
Unfortunately, we don't discover simple predictors (i.e., passage difficulty) of when annotators are likely to use heuristics.

A qualitative analysis of the collected data reveals that heuristic-seeking annotators are more likely to create examples that are not valid, and require simpler word-matching on explicitly stated information (\S\ref{sec:qual}). Crucially, this suggests that measurements of heuristic usage, such as those examined in this paper, can provide a general method to find unreliable examples in crowdsourced data, and direct our search for discovering artifacts in the data.
% To understand which parts of the annotation pipeline contribute to heuristic-seeking behavior in annotators, we also tease apart the effect of components inherent to the task (for eg, passage difficulty) as opposed to the annotators themselves (for eg, annotator fatigue) (\S\ref{sec:influencers}).
% Unfortunately, we discover few simple predictors (i.e. passage difficulty) of when annotators are likely to use heuristics.
Because we implicate heuristic use in terms of robustness and data quality, we suggest future dataset creators track similar features and evaluate model sensitivity to annotator heuristic use.\footnote{Our code and collected data is available at \href{https://github.com/chaitanyamalaviya/annotator-heuristics}{https://github.com/chaitanyamalaviya/annotator-heuristics}.}

\section{Background and Related Work}

\paragraph{Cognitive Heuristics.} The study of heuristics in human judgment, decision making, and reasoning is a popular and influential topic of research \cite{simon1956rational, tversky1974judgment}. Heuristics can be defined as mental shortcuts, that we use in everyday tasks for fast decision-making.
% These heuristics reduce the amount of cognitive effort that we expend on tasks \cite{shah2008heuristics} and provide fast and frugal ways to make decisions \cite{gigerenzer2011heuristic}. 
For example, \citet{tversky1974judgment} asked participants whether more English words begin with the letter \textit{K} or contain \textit{K} as the $3^{rd}$ letter, and more than 70\% participants chose the former because words that begin with \textit{K} are easier to recall, although that is incorrect. This is an example of the availability heuristic. Systematic use of such heuristics can lead to cognitive biases, which are irrational patterns in our thinking.

At first glance, it may seem that heuristics are always suboptimal, but previous work has argued that heuristics can lead to accurate inferences under uncertainty, compared to optimization \cite{gigerenzer2011heuristic}.
% On the other hand, \citet{shah2008heuristics} describe heuristics as a consequence of cognitive laziness.
We hypothesize that heuristics can play a considerable role in determining data quality and their impact depends on the exact nature of the heuristic. Previous work has shown that crowdworkers are susceptible to cognitive biases in a relevance judgement task \cite{eickhoff2018cognitive}, and has provided a checklist to combat these biases \cite{draws2021checklist}.
In contrast, our work focuses on how potential use of such heuristics can be measured in a writing task, and provides evidence that heuristic use is linked to model brittleness. 

Features of annotator behavior have previously been useful in estimating annotator task accuracies \cite{rzeszotarski2011instrumenting, goyal2018your}. Annotator identities have also been found to influence their annotations \cite{hube2019understanding, sap2022annotatorsWithAttitudes}. Our work builds on these results and estimates heuristic use with features to capture implicit clues about data quality.

\paragraph{Mitigating and discovering biases.}
The presence of artifacts or biases in datasets is well-documented in NLP, in tasks such as natural language inference, question answering and argument comprehension \cite[\textit{inter alia}]{gururangan2018annotation, mccoy2019right, niven-kao-2019-probing}. These artifacts allow models to solve NLP problems using unreliable shortcuts \cite{geirhos2020shortcut}. Several researchers have proposed approaches to achieve robustness against known biases.
% These include ensembling models with known bias models \cite{clark-etal-2019-dont, he-etal-2019-unlearn, karimi-mahabadi-etal-2020-end, utama-etal-2020-towards, clark-etal-2020-learning}, data augmentation \cite{min-etal-2020-syntactic, kaushik2020learning}, multi-task learning \cite{tu-etal-2020-empirical} and adversarial filtering \cite{le2020adversarial}.
We refer the reader to \citet{wang2022measure} for a comprehensive review of these methods. Targeting biases that are unknown continues to be a challenge, and our work can help find examples which are likely to contain artifacts, by identifying heuristic-seeking annotators.

Prior work has proposed methods to discover shortcuts using explanations 
of model predictions \cite{lertvittayakumjorn2021explanation}, including sample-based explanations \cite{han-etal-2020-explaining} and input feature attributions \cite{bastings2021will, pezeshkpour-etal-2022-combining}. Other techniques that can be helpful in diagnosing model biases include building a checklist of test cases \cite{ribeiro-etal-2020-beyond, ribeiro-lundberg-2022-adaptive}, constructing contrastive \cite{gardner-etal-2020-evaluating} or counterfactual \cite{wu-etal-2021-polyjuice} examples and statistical tests \cite{gururangan2018annotation, gardner-etal-2021-competency}. Our work is complementary to these approaches, as we provide an alternative approach to bias discovery that is tied to annotators.

\paragraph{Improved crowdsourcing.} 
A related line of work has studied modifications to crowdsourcing protocols to improve data quality \cite{bowman-etal-2020-new, nangia-etal-2021-ingredients}. In addition, model-in-the-loop crowdsourcing methods such as adversarial data collection \cite{nie-etal-2020-adversarial} and the use of generative models \cite{bartolo-etal-2022-models, liu2022wanli} have been shown to be helpful in creating more challenging examples. We believe that tracking annotator heuristics use can help make informed adjustments to crowdsourcing protocols.

\section{Annotation Protocol}

We consider multiple-choice reading comprehension as our crowdsourcing task, because of the richness of responses and interaction we can get from annotators, which allows us to explore a range of hypothetical heuristics. We describe here the methodology for our data collection.

We provided annotators on Amazon Mechanical Turk with passages and ask them to write a multiple-choice question with four options. We used the first paragraphs of ‘vital articles’ 
% (known to benchmark the quality of Wikipedia) 
from the English Wikipedia\footnote{Wikipedia Level 4 vital articles: \url{https://en.wikipedia.org/wiki/Wikipedia:Vital_articles/Level/4}}, and ensured that passages are at least 50 words long and at most 250 words long. Passages spanned 11 genres including arts, history, physical sciences, and others, and passages were randomly sampled from the 10K passages. Annotators were asked to write challenging questions that cannot be answered by reading just the question or passage alone, and have a single correct answer. Further, they were asked to ensure that passages provided sufficient information to answer the question while allowing questions to require basic inferences using commonsense or causality.

Annotators were first qualified to avoid spamming behavior. This qualification checked for spamming behavior in the form of invalid questions, and not example quality. Annotators were then asked to write a multiple-choice question to 4 passages in a single HIT on MTurk. Annotators were asked to not work on more than 8 HITs. We collected 1225 multiple-choice question-answer pairs from 73 annotators. In addition, we also logged their keystrokes and the time taken to complete an example (ensuring that time away from the screen was not counted). Our annotation interface was built upon \citet{nangia-etal-2021-ingredients}. For other details about our annotation protocol, please refer to Appendix~\ref{app:A}.

\section{Cognitive Heuristics in Crowdsourcing}
\label{sec:heuristics}
% describe cognitive heuristics and some seminal work in the field of heuristics + give examples of heuristics in daily life

\begin{table*}[t]
    \small
    \centering
    \begin{tabular}{p{0.3\linewidth}|p{0.6\linewidth}}
    \textbf{Consequence of cognitive heuristic}  & \textbf{Featurization} \\ \toprule
    Satisficing (lowtime)  &  (1) time, (2) log (time), (3) time / doc length, (4) log (time / doc length)\\
    Satisficing (loweffort)  & (1) question length, (2) keystroke length, (3) question+ops length, \\
    & (4) question+ops length / keystroke length  \\
    Availability (first option bias)  & First option is marked as correct answer \\
    Availability (serial position) &  Correct answer matches span in first or last sentence of passage  \\
    Representativeness (word overlap) &  Average word overlap in all pairs of examples by annotator\\
    Representativeness (copying) & (1) Length of longest common subsequence (lcs) b/w doc \& question, \\
    & (2) Max of normalized length of lcs between doc \& \{question, options\}, \\
    & (3) Normalized avg of length of lcs between doc \& \{question, options\} \\
    \bottomrule
    \end{tabular}
    \caption{Consequences of cognitive heuristics and featurizations for multiple-choice reading comprehension data.}
    \label{tab:heuristic}
\end{table*}

Cognitive heuristics are mental shortcuts, that humans employ in problem-solving tasks to make quick judgments \cite{simon1956rational, tversky1974judgment}. 
% At first glance, it may seem that heuristics are always suboptimal, but previous work has argued that heuristics can lead to accurate inferences under uncertainty, compared to optimization \cite{gigerenzer2011heuristic}. We are interested in exploring the precise relationship between these heuristics and resulting data quality. 
% Instead, we believe that the relationship between a heuristic's impact on data quality depends on the exact nature of the heuristic. We explore a range of cognitive heuristics that could be put to use for our task, and analyze their prevalence and impact on data quality. For a more comprehensive review of cognitive biases in crowdsourcing, we refer the readers to \citet{draws2021checklist}.
Annotators, tasked with authoring natural language examples, are not infallible to using such heuristics. 
% In fact, previous work has suggested that annotators can be susceptible to cognitive biases \cite{eickhoff2018cognitive, draws2021checklist}. 
We hypothesize that, in writing tasks, reliance on heuristics is a traceable indicator of poor data quality.
In this section, we identify several heuristics, their consequences in annotator behavior, and features to track them.
Later, we also show they are predictors of qualitatively important dimensions of data.
%Our experiments later show that features and heuristics we identify in the section below predict solvability by biased models.% and ourqualitative analysis.

\subsection{Methodology}

To test the above hypothesis, we consider several known cognitive heuristics which could be relevant for our task. This list is not comprehensive, and we refer the readers to prior work for a thorough overview of cognitive biases \cite{shah2008heuristics, draws2021checklist}. To tangibly measure the potential usage of a heuristic, we featurize each heuristic into a measurable quantity that can be computed automatically for an example (see Table~\ref{tab:heuristic}). While we do not conclusively determine that an annotator is using a heuristic, we explore various featurizations that align with the intuition behind each heuristic. These featurizations can sometimes be mapped to multiple heuristics that interact together, but for ease of presentation, we list them under the most related cognitive heuristic. These help us create \textit{\textbf{annotator heuristic \hname}}, which contain average heuristic values across all of an annotator's examples.
% we draw inspiration from prior work \cite{shah2008heuristics, draws2021checklist} on common cognitive biases that humans are subject to, especially during crowdsourcing, and form a list of heuristics which we think could be put to use for our task. 
% We use these feature values to create an \textit{\textbf{annotator heuristic profile}}, . We then flag those annotators who show strongest signs of heuristic use, as being users of a specific heuristic.

% \citet{simon1956rational} first began the study of human heuristics, describing a formal model for how humans make decisions under uncertainty. \citet{tversky1974judgment} discuss various judgmental heuristics such as the availability heuristic and and how they result in cognitive biases.

To verify if our instantiation of a heuristic aligns with heuristic-seeking tendencies in annotators, we measure correlations of heuristic values with annotator performances on a battery of psychological tests \cite{frederick2005cognitive, toplak2014assessing, sirota2021measuring}, described in \S\ref{sec:crt}. %Below we describe the heuristics we analyze in our study.

\subsection{Heuristics Studied}

\paragraph{Satisficing:} Satisficing is a cognitive heuristic that involves making a satisfactory choice, rather than an optimal one \cite{simon1956rational}. In terms of  mental process, strong satisficing can involve inattention to information and lack of information synthesis. In social cognition, \citet{krosnick1991response} described how satisficing can manifest in various patterns in survey responses. For example, survey-takers might pick the same response to several questions in sequence, pick a random response, or use the acquiescence bias (where they always choose to \textit{agree} with the given statement). A potential outcome of satisficing in our task is low time spent on the task and low effort put into forming a question. 

% An annotator who is working on tasks hurriedly might be using heuristics to frame questions. On the other hand, if a worker is spending ample amount of time on a task, they might be more likely to create a difficult question. 
Assuming the working time is $t$ and number of tokens in a passage $d$ is $l_d$, we consider the following \textit{lowtime} featurizations: (1) $t$, (2) $\log{t}$ , (3) $t/l_d$, (4) $\log(t/l_d)$.\footnote{Previous work shows that taking the logarithm normalizes the response time distribution \cite{whelan2008effective}.}

We estimate an annotator's amount of effort through their responses. An annotator who is consistently editing their work or writing long questions might be attempting to thoughtfully draft their question. While this may not always be true (for eg, a worker might spend time thinking about their question and only start writing later), we hypothesize that often, short responses can be indicators of satisficing. Given the number of words found in a stream of keystrokes, $k$, the question $q$, and all options $o_i$ is $l_k$, $l_q$ and $l_o$, we consider these \textit{loweffort} featurizations: (1) $l_q$, (2) $l_k$, (3) $l_q+l_o$, (4) $(l_q+l_o)/l_k$.

\paragraph{Availability heuristic:} The tendency to rely upon information that is more readily retrievable from our memory is the availability heuristic \cite{tversky1973availability}. For example, after hearing about a plane crash on the news, people may overstate the dangers of flying. For our task, once an annotator has read a passage and formulated a question, the question and the correct answer are likely to be readily available in their mind. This could cause them to write that information before any of the distractor options. Therefore, we check whether the first option specified for an example is also the correct answer (\textit{first option bias}).

Another consequence of this heuristic is the serial-position effect. When presented with a series of items like a list of words or items in a grocery list, people recall the first and last few items from the series better than the middle ones \cite{murdock1962serial, ebbinghaus1964memory} because of their easier availability. This effect can also be explained as a combination of the primacy effect and recency effect. To test if an annotator anchors their questions on the first or last sentence of the passage due to this heuristic, we check if the correct answer marked for an example matches a span in the first or last sentence of the passage (\textit{serial position}).

\paragraph{Representativeness heuristic:} The representativeness heuristic is our tendency to use the similarity of items to make decisions \cite{kahneman1972subjective}. For example, if a person is picking a movie to watch, they might think of movies they previously liked and look for those attributes in a new movie. Similarly, an annotator may repeat the same construction in their questions to ease decision-making (e.g., "which of the following is true?" or "what year did [event] happen?"). This could either mean that they are not fully engaged, or, they found a writing strategy that works well and they choose to stick to it. We measure this tendency by computing the average \textit{word overlap} across all pairs of questions from an annotator.

A different manner in which this heuristic can manifest is using similarity with the provided context, i.e., through \textit{copying}. Copying, or imitation, is a common building block that guides human behavior and decision making. In deciding what clothes to buy or which book to read, humans use imitation-of-the-majority to make quicker inferences with lesser cognitive effort \cite{garcia2009does, gigerenzer2011heuristic}. Similarly, annotators can have tendencies to copy text word-for-word from the context they are primed with, to reduce their cognitive load. Assuming \textsc{lcs} is a function that computes the length of the longest common subsequence between two sequences, we consider these featurizations for copying: (1) \textsc{lcs}$(d,q)$, (2) $\textnormal{max}(\textsc{lcs}(d,q), \textsc{lcs}(d,o))$ and (3) $\textnormal{avg}(\textsc{lcs}(d,q), \textsc{lcs}(d,o))$.

\subsection{Annotator Heuristic Traces}
% Examples level features of annotator behavior are aggregated to form heuristics statistics for annotators across the whole task. 
The consequences of heuristics we compute, as summarized in Table~\ref{tab:heuristic}, may not in themselves be problematic per example.
However, we claim that annotators who consistently rely on such heuristics may impart larger, harder-to-detect, undesirable regularities in data.

Annotator heuristic traces capture global behavioral trends per annotator.
For each annotator and heuristic, we average the heuristic values across all of the annotator's examples, 
%to construct our \hname, 
forming a matrix of annotators and their average heuristic values.

%These form a matrix of annotators with their average heuristic values.

\paragraph{Principal components of heuristics:} We also evaluate if a low-dimensional representation of an annotator's heuristic trace is useful for predicting data quality.
We compute the first principal component of this matrix to simultaneously consider multiple heuristic indicators.
% Then, we represent the annotators use of heuristics by computing the projection of their profile on the first principal component of heuristic use. 

\subsection{Cognitive Reflection Test}
\label{sec:crt}
\begin{figure}[t!]
    \centering
    \includegraphics[width=\columnwidth,height=9cm,keepaspectratio]{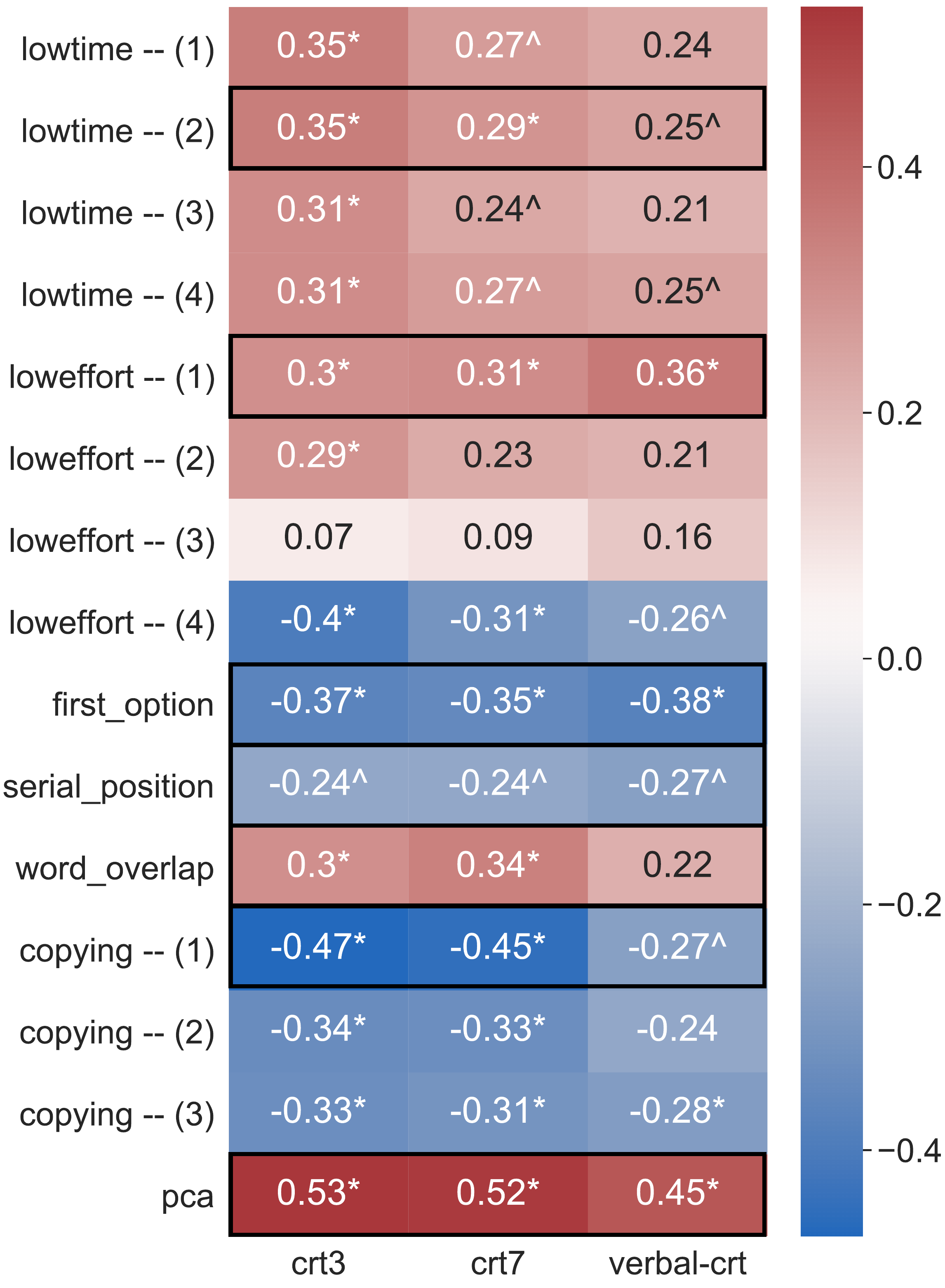}
    \caption{Correlations of annotator scores on the CRT and their average features values for each heuristic. Feature names, left, correspond to feature names from Table 1. The CRT3 includes the original questions from \citet{frederick2005cognitive} and CRT7 includes 4 more questions from \citet{toplak2014assessing}. The black boxes indicate the featurization with the highest average correlations for the heuristic. $*$ indicates $p<0.01$ and \string^ indicates $p<0.1$.}
    \label{fig:crt}
\end{figure}

Although we cannot determine whether an annotator is definitively using a heuristic, we can probe if our features correlate with heuristic-seeking tendencies in annotators. Previous work in cognitive psychology has designed tests measuring such tendencies. These help us validate the psychological plausibility of our features, ensuring they are generally applicable. 

Perhaps the best known test of heuristic-seeking tendencies is the Cognitive Reflection Test (CRT) \cite{frederick2005cognitive}. The test has 3 questions, but we instead use the 7-item CRT from \citet{toplak2014assessing} to find more variance among annotators. 
The numerical CRT requires mathematical reasoning and previous work has highlighted that its results might be conflated with mathematical reasoning capabilities. 
Further, since our task requires writing, we also perform the verbal CRT \cite{sirota2021measuring}. 
This test has 9 items\footnote{We exclude a question that requires cultural knowledge.}, and is known to correlate well with the numerical CRT, and other indicators of cognitive capabilities. 
The questions in these tests are provided in Appendix~\ref{app:B}.
\footnote{The use of the CRT has issues due to repeated exposure \cite{stieger2016limitation, haigh2016has}, so we ensured that the names/quantities are different from the ones used in the original questions. We also emphasize that the CRT does not provide interpretability into the mechanism of the heuristics, whereas our individual heuristic features do.}

We asked annotators who completed at least 5 question writing examples to do two surveys asking logical questions (CRT-7 and Verbal CRT). 49 of 59 annotators completed the surveys.
We then compute Pearson correlations between annotator accuracies on the three versions of the CRT, and values in their heuristic traces, %the averaged heuristic feature values across all of an annotator's examples.
%These correlations are 
shown in Figure~\ref{fig:crt}.
The results indicate that our featurizations have significant, medium correlations with the CRTs, and the PCA projection, which captures multiple heuristics, has the highest correlations.
%Notably, the word overlap feature is positively correlated with the CRT, suggesting that high similarity in questions is predictive of higher CRT scores.
For the sake of further analysis, for each feature group, we use the feature that has the highest average correlations with the CRT tests (enclosed in black boxes in Figure~\ref{fig:crt}).

\begin{figure*}[h!]
    \centering
    \includegraphics[width=\textwidth]{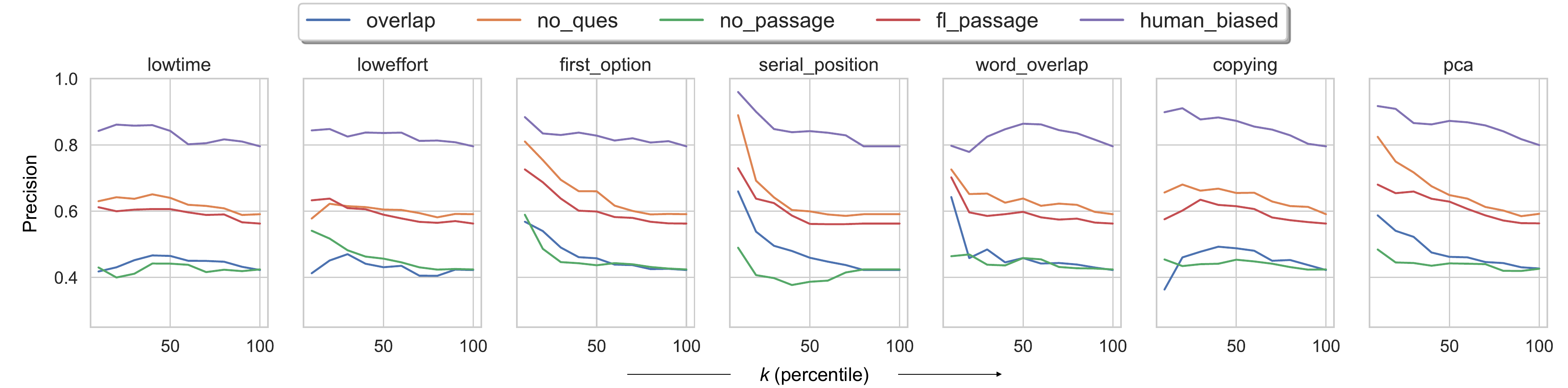}
    \caption{Precision of labeling heuristic examples $H_k$ as solvable by biased models, when the set $H_k$ is formed by examples from the $k^{th}$ percentile of heuristic-seeking annotators.}
    \vspace{-10pt}
    \label{fig:perc}
\end{figure*}
\begin{figure*}[h!]
    \centering
    \includegraphics[width=\textwidth]{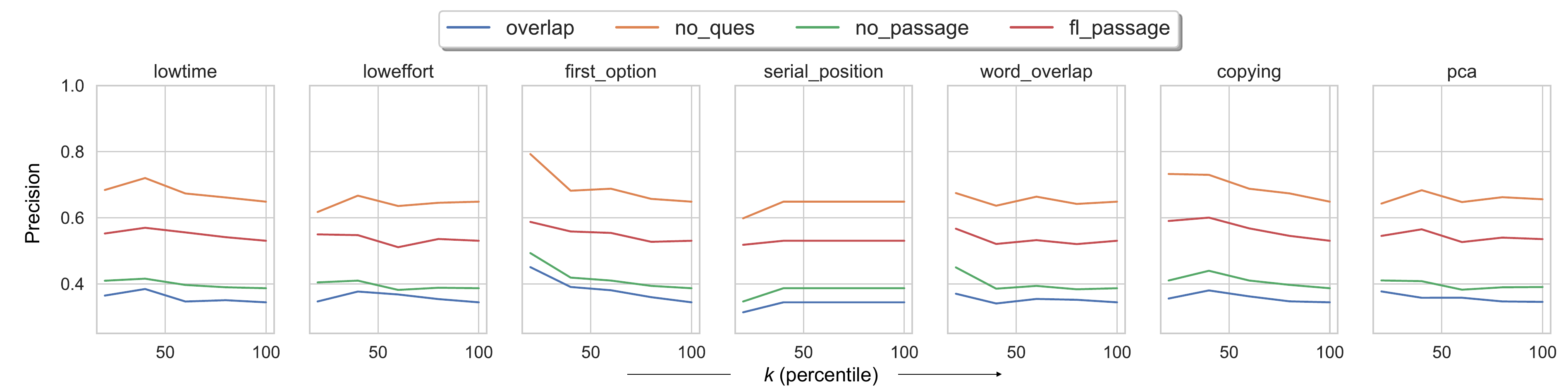}
    \caption{Precision of labeling heuristic examples $H_k$ as solvable by biased models %, when the set $H_k$ is formed by examples from the $k^{th}$ percentile of heuristic-seeking annotators. 
    on data from \citet{sugawara2022makes}.}
    \label{fig:perc2}
\end{figure*}

\section{Biased Model Solvability}
\label{sec:bias}

Annotator heuristic traces are cognitively plausible measures that we hypothesize are indicators of large, potentially undesirable patterns annotators impart on data. 
To verify this, we test if examples created by heuristic-seeking annotators are more easily solvable by biased models. 

We consider heuristic examples as examples from those annotators who score highly on our heuristic indicator features. Given the initial set of examples $D$, we distinguish a subset as heuristic, $H_k$, formed by all examples from annotators in the top $k$\% of average heuristic use across all annotators. We form such a subset independently for all heuristic indicator features we consider.\footnote{We exclude those annotators who wrote less than 5 examples and exclude all invalid examples.}
When $H_k$ is formed from the top quartile (\textit{k}=25), 68\% of annotators have examples included in at least 1 heuristic set, and 14\% in for at least 4/6 heuristic sets. We find that few annotators never use heuristics. 

% , by considering the top quartile of annotators with the most extreme average heuristic values. For instance, we label examples from the annotators who take the least amount of time to write examples, as ones that have been created using heuristics.

Next, we evaluate how well biased models perform on heuristic subsets ($H_k$) compared to the remaining examples, $D \setminus H_k$. We evaluate a few biased models, trained to use unreliable heuristics, on examples created with or without heuristics. Below we describe the biased models we use. In all cases, we train or finetune models on QA data from \citet{nangia-etal-2021-ingredients} and evaluate them on our data. For hyperparameter settings, please see Appendix~\ref{app:C}.

\paragraph{Lexical Overlap Model (overlap).} We train a logistic regression classifier by building upon features from the bias-only model from \citet{clark-etal-2019-dont}. Assuming the concatenated passage and question are the context for each option, we use the following features: 1) is the option a subsequence of context, 2) do all words in the option exist in context, 3) the fraction of words in the option that exist in context, 4) the log of length difference between the context and the option, 5) the average and maximum of minimum distance between each context word with each option word using 300-dimensional fastText embeddings \cite{joulin2016bag}. We then pick the option with the highest probability as the model prediction. The model achieves an accuracy of 42.27\% on $D$.

\paragraph{Partial Input Models.} As a benchmark for diagnosing the collected data, we consider several partial input models. These include no passage (\textbf{no\_passage}), no question (\textbf{no\_ques}), first \& last sentence of passage only (\textbf{fl\_passage}). We use a RoBERTa-Large \cite{liu2019roberta} model initially finetuned on RACE \cite{lai-etal-2017-race} and further trained on the baseline data from \citet{nangia-etal-2021-ingredients}. These models achieve accuracies of 42.44\%, 59.49\%, and 55.98\% on $D$, respectively, demonstrating better than random performance.

\paragraph{Human heuristic solvability (human\_biased).} In addition to biased models, we also consider an implicit notion of example difficulty from a biased human. Specifically, we evaluate whether a human can answer an example just by skimming the passage. We use an interface where a passage is only visible for 30 seconds, after which, a human needs to answer the question. One of the authors conducted this annotation for the collected examples and achieved an accuracy of 79.79\%.

% \paragraph{Setup.}

% \begin{figure}[h!]
%     \centering
%     \includegraphics[width=0.45\textwidth,keepaspectratio=True]{images/models_quartile.png}
%     \caption{Accuracies of examples created with or without heuristics. These plots assume that the top quartile of heuristic-seeking annotators write examples where heuristic=1. The dotted line shows the accuracy of the biased model on all the examples.}
%     \label{fig:models}
% \end{figure}

\paragraph{Results.} 
%  In our analysis, we labeled examples from those annotators who belonged to the top quartile of strong heuristic users. 
 Figure~\ref{fig:perc} shows the precision of $H_k$ being solvable by biased models, as $k$ is varied. As we can see from the plots, there is a downward trend as the percentile is increased for all heuristics. The features for the availability and representative heuristic, and the PCA projection are particularly effective. This suggests that strongly heuristic-seeking annotators are more likely to create examples solvable by biased models.
 
 \paragraph{Other non-Wikipedia domains.} To test if the heuristics we considered are indicative of solvability by biased models in domains other than Wikipedia, we repeated our analysis on 1,982 examples from the standard data collection setting in \citet{sugawara2022makes}, who collected questions for passages from many different sources. 
 %There were only 18 workers who completed at least 5 examples, which makes this analysis difficult.
 The precision plot is shown in Figure~\ref{fig:perc2}. With the exception of serial-position, heuristic-seeking features identify annotators that create examples more easily solvable by biased models in these domains too.\footnote{The serial-position feature might be more effective for Wikipedia because they tend to be more factual, which makes it easier to form questions using the first or last sentence.}

% We plot the accuracy of biased models on examples created from the set H and D-H in Figure~\ref{fig:models}. Generally, we find that strong heuristic use among annotators is predictive of solvability by biased models. Examples created with copying, representativeness, low effort or availability, are more easily solvable by biased models (from 3\% up to 16\% difference) and a biased human (from 3\% up to 13\% difference). All biased models except \textit{no passage} have higher accuracy on examples created with low time, and the serial-position heuristic.

\begin{figure}[t!]
    \centering
    \includegraphics[width=\columnwidth,height=7cm,keepaspectratio]{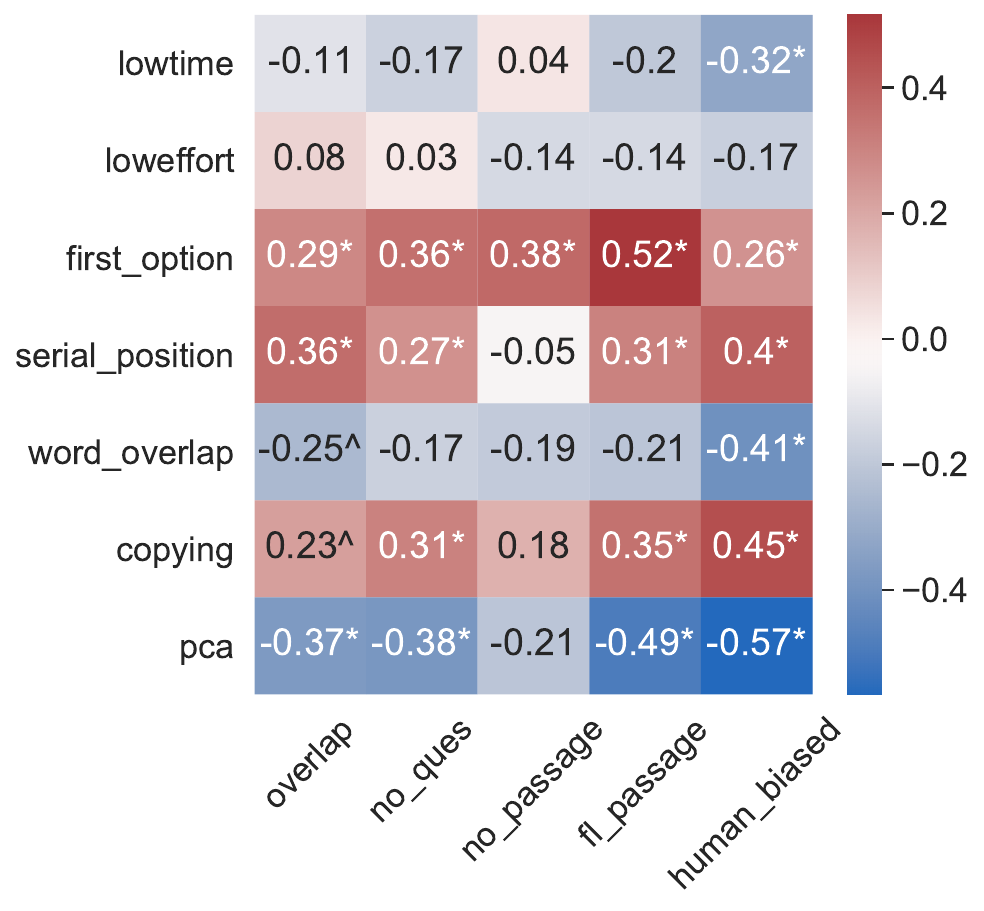}
    \caption{Pearson correlations of annotators' average heuristic values and accuracies with biased models on their annotated examples. $*$ and \string^ indicate $p<0.01$ and $p<0.1$.}
    \label{fig:overall_correls}
\end{figure}

\paragraph{As a predictor of bias.} In addition to evaluating the predictiveness of annotator heuristic features at the extreme, we also evaluated whether heuristic features are predictive of solvability by biased models across annotators. Specifically, we calculated Pearson correlations between annotators' average heuristic values, and the accuracies of biased models on their examples, in Figure~\ref{fig:overall_correls}. These correlations are not strong for the satisficing heuristics, but we do notice some significant, medium correlations for the other heuristics we studied. Importantly, we contrast this with the same correlations measured over the entire pool of data (without averaging per annotator). Those correlations, shown in Figure~\ref{fig:overall_correls_pooled} in the Appendix, are much weaker showing the value of our annotator-level measures. % for identifying problematic examples.
%the importance of studying annotators' personal statistics for identifying problematic examples. 

\paragraph{Model generalization across annotators.} 
Previous work showed that models do not generalize well to annotator-based random splits of crowdsourced datasets, suggesting models might be learning annotator-specific biases \cite{geva-etal-2019-modeling}.
We suspect that generalization might deteriorate when models are trained on heuristic-seeking annotators, as models could more easily specialize to their examples. Hence, we ask whether heuristic-based splits (\textbf{heuristic}) lead to worse performance than random annotator splits (\textbf{random}).

While controlling the number of training examples, we trained models on examples from heuristic-seeking annotators or a random set of annotators, and test on the remaining examples. For heuristic-based splits, we train on examples in $H_{33}$, the top 33\% of heuristic-seeking annotators for a heuristic indicator.
For random annotator splits, we resampled splits with 3 random seeds and report means. In addition, we trained models on random splits of the same training size (\textbf{random-pooled}), where data is not split by annotator. The accuracies on these splits are shown in Table~\ref{tab:annotator_splits}. We find generalization is poorer for almost all of the heuristic-based annotator splits compared to random annotator splits. 
This suggests that heuristic-based splits can serve as natural challenge sets and inadvertently sampling heuristic-seeking annotators for training may not generalize well. % to more complex data distributions.
% in circumstances that require less heuristic reasoning. 

% Specifically, we train logistic regression models to predict whether our biased models answer an example correctly. To train these models, we use heuristic features as well as other annotation metadata such as annotator ID, document length, etc (listed in Appendix~\ref{app:C}). The classifiers are able to achieve accuracies above a majority-class classifier (results in Table\ref{tobeadded}), which suggests that heuristic features are predictive of several known biases. We also show correlations between heuristic features and biased model solvability in Appendix\ref{app:E} \chaitanya{To add}.

\begin{table}[t!]
\small
\centering
{
\setlength{\tabcolsep}{0.3em}
    \begin{tabular}{c|c|c|c|c}
    \textbf{Heuristic}  & heuristic & random & random-pooled & n \\
    \hline
    lowtime & $84.27$ & $87.42_{\pm 1.17}$ & $87.14_{\pm 0.64}$ & 394 \\
    loweffort & $85.06$ & $88.56_{\pm 1.01}$ & $86.27_{\pm 1.75}$ & 415 \\
    first\_option & $87.33$ & $87.92_{\pm 0.85}$ & $88.12_{\pm 1.63}$ & 341 \\
    serial\_position & $84.78$ & $88.07_{\pm 0.84}$ & $88.43_{\pm 0.6}$ & 389 \\
    word\_overlap & $86.86$ & $86.64_{\pm 2.16}$ & $87.50_{\pm 0.67}$ & 362 \\
    copying & $83.75$ & $86.94_{\pm 1.78}$ & $87.47_{\pm 1.02}$ & 332 \\
    pca & $83.40$ & $86.94_{\pm 0.77}$ & $87.52_{\pm 0.81}$ & 385 \\
    \hline
    \end{tabular}
}
\caption{Performance on heuristic-based, random annotator splits and random splits with the same training set size. We performed 3 runs on the randomly sampled splits, and report means and standard deviations.}
\label{tab:annotator_splits}
\end{table}

\begin{figure*}
    \centering
    \includegraphics[width=\textwidth]{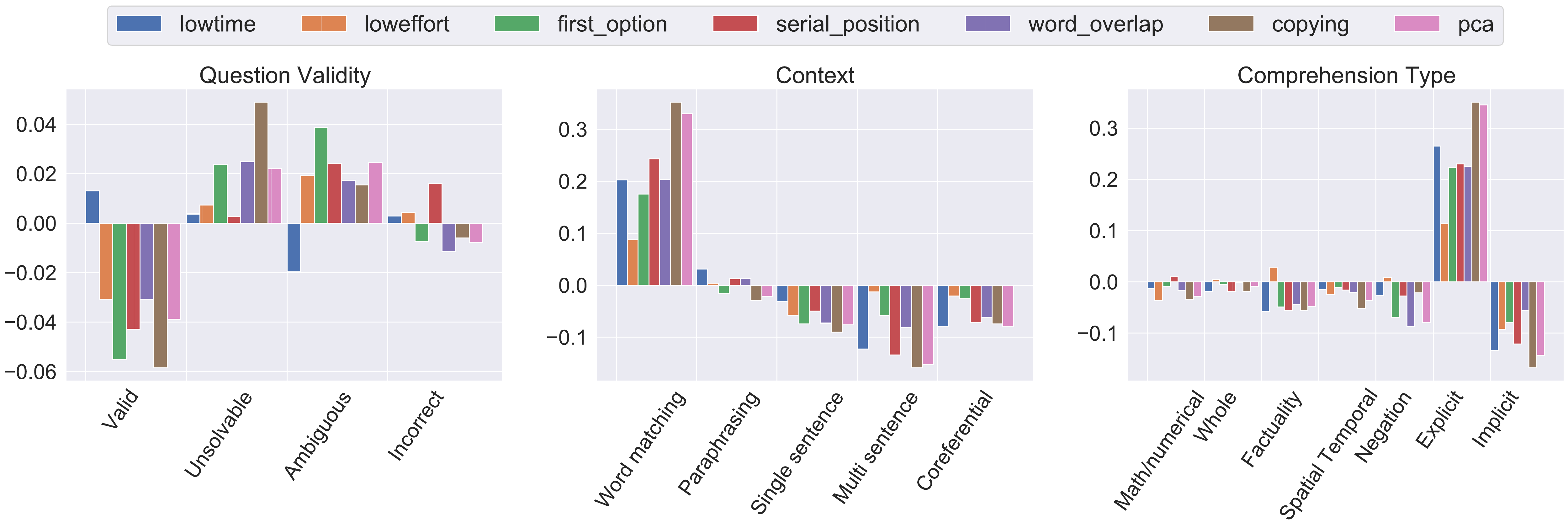}
    \caption{Percentage difference of examples in heuristic set, $H_{25}$, and the remaining examples, $D \setminus H_{25}$, labeled as having a qualitative property. Examples in the heuristic set are less valid \& require more word matching based on explicitly stated information.}
    \label{fig:ques_annot}
\end{figure*}

\section{Influencers of heuristic behavior}
\label{sec:influencers}

Next, we aim to understand what role the annotation pipeline plays in influencing heuristic use among annotators. Various factors have been shown to determine example quality in crowdsourcing. These include task difficulty, incentives, annotator ability, motivation and fatigue \cite{krosnick1991response,yan2010modeling}. 
%We aim to understand what role some of these factors play in our task. Specifically, 
We looked at how such markers influence heuristic use among annotators.

% \paragraph{Task properties.}

We considered two types of measures that could indicate difficulty: \textit{passage length} (number of tokens) and inverse \textit{entity} count (doc length / number of named entities) in the passage. Longer documents, with fewer named entities, might provide context that is harder to form questions about.
Further, having completed more examples could make an annotator fatigued and/or gain expertise at the task. Hence, we also used the sequence \textit{index} of each example for an annotator.
We computed Pearson correlations between these indicators and the heuristic values for each annotator, and averaged the correlations across annotators.
Our results are summarized in Table~\ref{tab:correl_influencer}.
We find that neither of these factors show significant correlations with heuristic features among annotators.

\begin{table}[t!]
\small
\centering
{
\setlength{\tabcolsep}{0.3em}
    \begin{tabular}{c|c|c|c}
    \textbf{Heuristic}  & passage length & entity & index \\ %&$\rho (passage\_len)$ & $\rho (entity)$ & $\rho (index)$ \\
    \hline
    lowtime & 0.12 & 0.16 & -0.09\\
    loweffort & 0.09 & -0.03 & -0.01\\
    first\_option & 0.08 & 0.00 & 0.02\\
    serial\_position & -0.09 & -0.05 & -0.03\\
    copying & 0.13 & 0.06 & -0.05\\
    \hline
    \end{tabular}
}
\caption{Correlations between heuristic values and factors, averaged across annotators.}
\label{tab:correl_influencer}
\end{table}

% \paragraph{Annotator properties.}

% \begin{figure*}[h!]
%     \centering
%     \includegraphics[width=\textwidth]{images/percentile_sugawara.pdf}
%     \caption{Precision of labeling heuristic examples $H_k$ as solvable by biased models, when the set $H_k$ is formed by examples from the $k^{th}$ percentile of heuristic-seeking annotators. Based on data from \citet{sugawara2022makes}.}
%     \label{fig:perc2}
% \end{figure*}

\section{Qualitative Analysis}
\label{sec:qual}
%Next, we analyze if examples created by heuristic-seeking annotators are qualitatively different.
To better understand the differences in data produced by heuristic-seeking annotators, and otherwise, we conducted a comprehensive qualitative analysis of all our data.
%questions with qualitative properties that are considered desirable for creating challenging QA datasets. 
We annotate questions with properties inspired from previous work \cite{lai-etal-2017-race, trischler-etal-2017-newsqa, sugawara-etal-2018-makes, sugawara2022makes} along the following dimensions: \textbf{validity} (is the question answerable given the context in the passage), \textbf{context} (how much context from the passage is needed to answer the question), and \textbf{comprehension type} (what kinds of comprehension are needed to answer the question). Each question can have multiple labels.
%for context and comprehension type.
For a detailed description of these labels, please refer to Appendix~\ref{app:D}.

\paragraph{Results.} Figure~\ref{fig:ques_annot} presents the results of our annotation. We show the differences in the percentage of examples in the heuristic set, $H_{25}$, and the remaining examples, $D \setminus H_{25}$. 
First, examples in the non-heuristic set are more likely to be valid, and less likely to be unsolvable compared to the heuristic set. 
Further, we find that examples in the heuristic set often require simple word matching and paraphrasing, while the ones in the non-heuristic set, are more likely to require multi-sentence reasoning.
In terms of comprehension type, we find that heuristic examples are more likely to be answerable using information explicitly stated in the passage.
On the other hand, non-heuristic examples are more likely to require implicit inference.
These results suggest there are significant qualitative differences in examples from heuristic-seeking annotators.

\section{Discussion}

Our work measures the implications of annotators' potential use of cognitive heuristics in data quality.
% While many of these implications appear to degrade date quality, we believe this judgement should be left to practical applications that use the data.
%Instead, we hoped to better understand the underlying competencies data could impart to a model.  
% Instead, our goal was to understand the relationship between annotator heuristics, data, and models.
The analyses we present suggest that models are indirectly influenced by heuristic use and that previous observations, such as the success of partial input models, is a consequence. 
While many such consequences of heuristic use appear to be negative, we believe that this judgement should be left up to practical applications that use the data.
We propose a fruitful direction for characterizing what models learn from data by considering annotator behaviors.

%We find annotator use of cognitive heuristics predictive of the quality of data they produce, in terms of biased model solvability, model generalizability and qualitative analysis. 
Practically, it is an open question as to how we can control downstream data using annotator heuristic traces.
Instead, we propose that future annotation efforts minimally track indicators of heuristic usage, using task-specific features, in an effort to document how they are reflected in the collected data and trained models. This would entail releasing annotator-level labels for crowdsourced data, and releasing annotation metadata \cite{plank2022problem}.

\section{Limitations} One limitation of our study is that we analyze the implications of heuristic-seeking behavior in annotators for one task. Future work could consider extending this methodology to many annotation tasks. For example, in sentence-pair classification tasks such as textual entailment, or in annotation of machine translation or summarization datasets, annotator heuristics could be useful in determining the quality of data and the biases embedded in them. To find stronger signals in the annotator heuristic \hname, future work could consider training models to featurize heuristics.

\section*{Acknowledgements}

This work is supported in part by an Allen Institute for AI Young Investigator award and National Science Foundation grant SES-1847794.
We thank the anonymous reviewers, and the annotators who participated in our study. We also thank Dan Roth, Jacob Eisenstein and the University of Pennsylvania NLP group for helpful discussions.

\bibliography{anthology,custom}
\bibliographystyle{acl_natbib}

\appendix

\clearpage
\section{Crowdsourcing setup}
\label{app:A}

\begin{figure*}
    \centering
    \includegraphics[width=\textwidth]{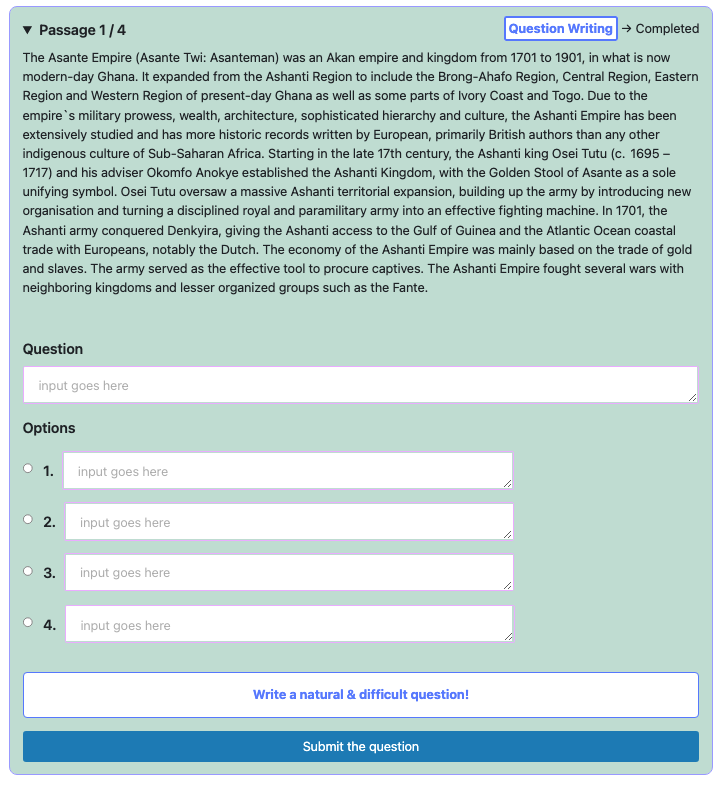}
    \caption{Annotation interface used for data collection.}
    \label{fig:interface}
\end{figure*}

For annotators to participate in our task, they needed to have an acceptance rate greater than or equal to 98\% and have at least 1000 approved HITs.
In addition, we required annotators to be located in US, UK or Canada. 
We estimated each HIT to take approximately 15 minutes, and paid \$4 per HIT (\$15 / hr). Figure~\ref{fig:interface} shows the interface presented to the annotators for data collection.

\section{Cognitive Reflection Tests}
\label{app:B}

We list the questions used in the numerical CRT and the verbal CRT in Table \ref{tab:numerical_crt} and Table \ref{tab:verbal_crt} respectively. The first 3 questions in Table \ref{tab:numerical_crt} correspond to the original CRT from \citet{frederick2005cognitive}.

\begin{table*}[t]
    \centering
    \small
    \begin{tabular}{p{0.7\linewidth}|p{0.1\linewidth}|p{0.1\linewidth}}
    \textbf{Question}  & \textbf{Intuitive Answer}  & \textbf{Correct Answer} \\ \toprule
    1) A carpet and a lamp cost \$450 in total. The carpet costs \$400 more than the lamp. How much does the lamp cost?
    &  \$50    &   \$25   \\
    2) It takes 10 computers 10 minutes to run 10 programs. How many minutes does it take 500 computers to run 500 programs? 
    &  500    &  10   \\
    3) There is a patch of lily pads in a pond. The patch doubles in size every day. If it takes 100 days for the patch to cover the entire pond, how many days would it take to cover half the pond?
    &  50    &  99   \\
    4) If Jason can drink one barrel of water in 6 days, and Jen can drink one barrel of water in 12 days, how long would it take them to drink one barrel of water together?
    &  9    &  4   \\
    5) Aidan received both the 25th highest and the 25th lowest mark in the class. How many students are in the class?
    &  50    &  49   \\
    6) A farmer buys a sheep for \$500, sells it for \$600, buys it back for \$700, and sells it finally for \$800. How much has he made?
    &  \$100    & \$200   \\
    7) Ramona decided to invest \$5,000 in the stock market early in 2008. Six months after she invested, on July 17, the stocks she had purchased were down 50\%. Fortunately for Ramona, from July 17 to October 17, the stocks she had purchased went up 75\%. At this point, does Ramona have a) the same amount of money as when she invested, b) more money, c) less money ? (Respond with a, b or c)  &  b    &  c \\
    \bottomrule
    \end{tabular}
    \caption{Questions in the numerical CRT.}
    \label{tab:numerical_crt}
\end{table*}

\begin{table*}[t]
    \centering
    \small
    \begin{tabular}{p{0.7\linewidth}|p{0.1\linewidth}|p{0.15\linewidth}}
    \textbf{Question}  & \textbf{Intuitive Answer}  & \textbf{Correct Answer} \\ \toprule
    1) Angie's father has 5 daughters but no sons—Nana, Nene, Nini, Nono. What is the fifth daughter's name probably?
    & Nunu   &   Angie  \\
    2) If you were running a race, and you passed the person in 5th place, what place would you be in now?
    &  4th    &  5th   \\
    3) It is a stormy night and a plane takes off from JFK airport in New York. The storm worsens, and the plane crashes - half lands in the United States, the other half lands in Canada. In which country do you bury the survivors?
    &  USA    &  we do not bury survivors   \\
    4) A monkey, a squirrel, and a bird are racing to the top of a coconut tree. Who will get the banana first, the monkey, the squirrel, or the bird?
    &  bird    &  there is no banana on a coconut tree   \\
    5) In a one-storey pink house, there was a pink person, a pink cat, a pink fish, a pink computer, a pink chair, a pink table, a pink telephone, a pink shower—everything was pink! What colour were the stairs probably?
    &  pink    &  no stairs in a one-storey house   \\
    6) The wind blows west. An electric train runs east. In which cardinal direction does the smoke from the locomotive blow?
    & west   & no smoke from an electric train   \\
    7) If you have only one match and you walk into a dark room where there is an oil lamp, a newspaper and wood— which thing would you light first?  &  oil lamp &  match \\
    8) Would it be ethical for a man to marry the sister of his widow?  &  no &  not possible \\
    9) Which sentence is correct: (a) ‘the yolk of the egg are white’ or (b) ‘the yolk of the egg is white’? &  b &  the yolk is yellow \\
    \bottomrule
    \end{tabular}
    \caption{Questions in the verbal CRT.}
    \label{tab:verbal_crt}
\end{table*}

\section{Hyperparameter Settings}
\label{app:C}
\paragraph{Lexical overlap model.} The logistic regression was trained with C=100 and a maximum of 100 iterations for convergence with the scikit-learn library \cite{scikit-learn}.

\paragraph{Partial input models.} The partial input models were trained with a learning rate of 1e-4 and batch size of 1, for 4 epochs and all the default hyperparameters in the multiple-choice QA example in the Transformers library \cite{wolf2019huggingface}. These experiments took approximately a week of compute time on a single Quadro RTX 6000 GPU.

\section{Correlations with pooled data}

\begin{figure}[t]
    \centering
    \includegraphics[width=\columnwidth,height=7cm,keepaspectratio]{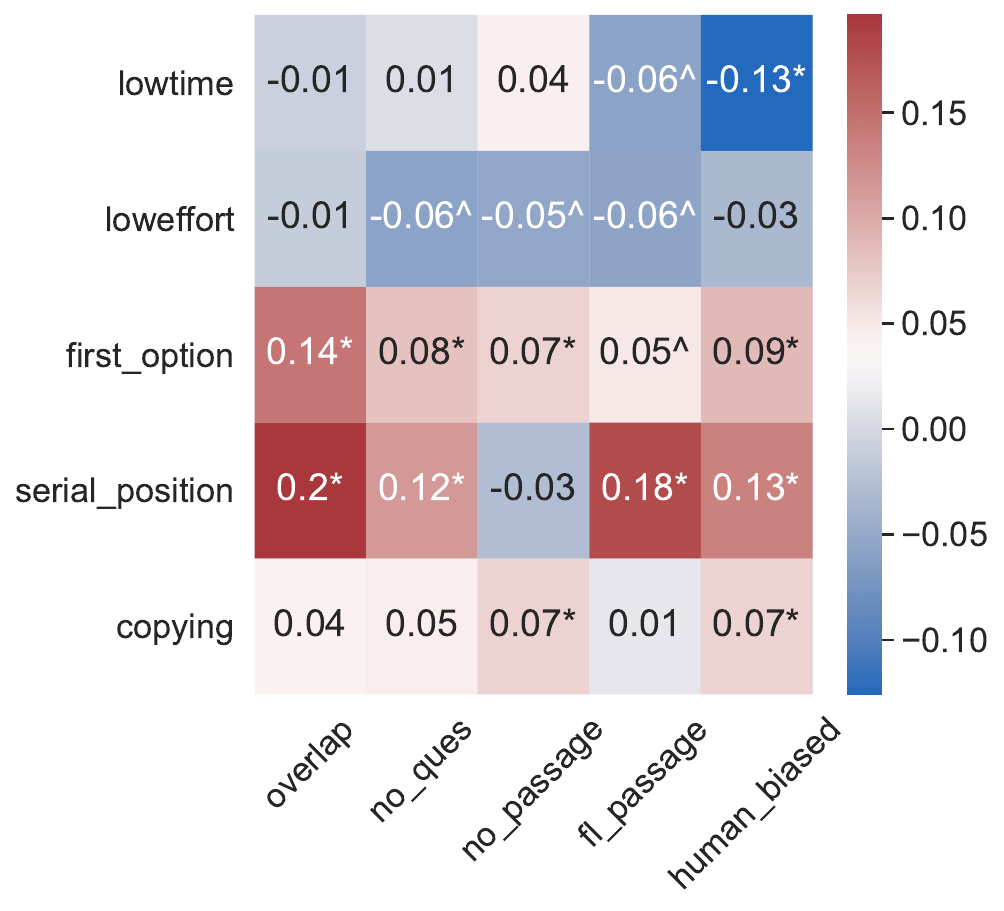}
    \caption{Pearson correlations of average heuristic values and biased model solvability, pooled across all examples. We exclude \textit{word overlap} since it is computed across all examples of an annotator and is not a sample-level measure. $*$ indicates $p<0.01$ and \string^ indicates $p<0.1$.}
    \label{fig:overall_correls_pooled}
\end{figure}

In Figure~\ref{fig:overall_correls_pooled}, we show correlations between heuristic features and biased model accuracies when all examples are pooled together. We contrast this with the annotator-wise plots shown in Figure~\ref{fig:overall_correls}.

\section{Question Annotation Scheme}
\label{app:D}

\begin{figure*}
    \centering
    \includegraphics[width=\textwidth]{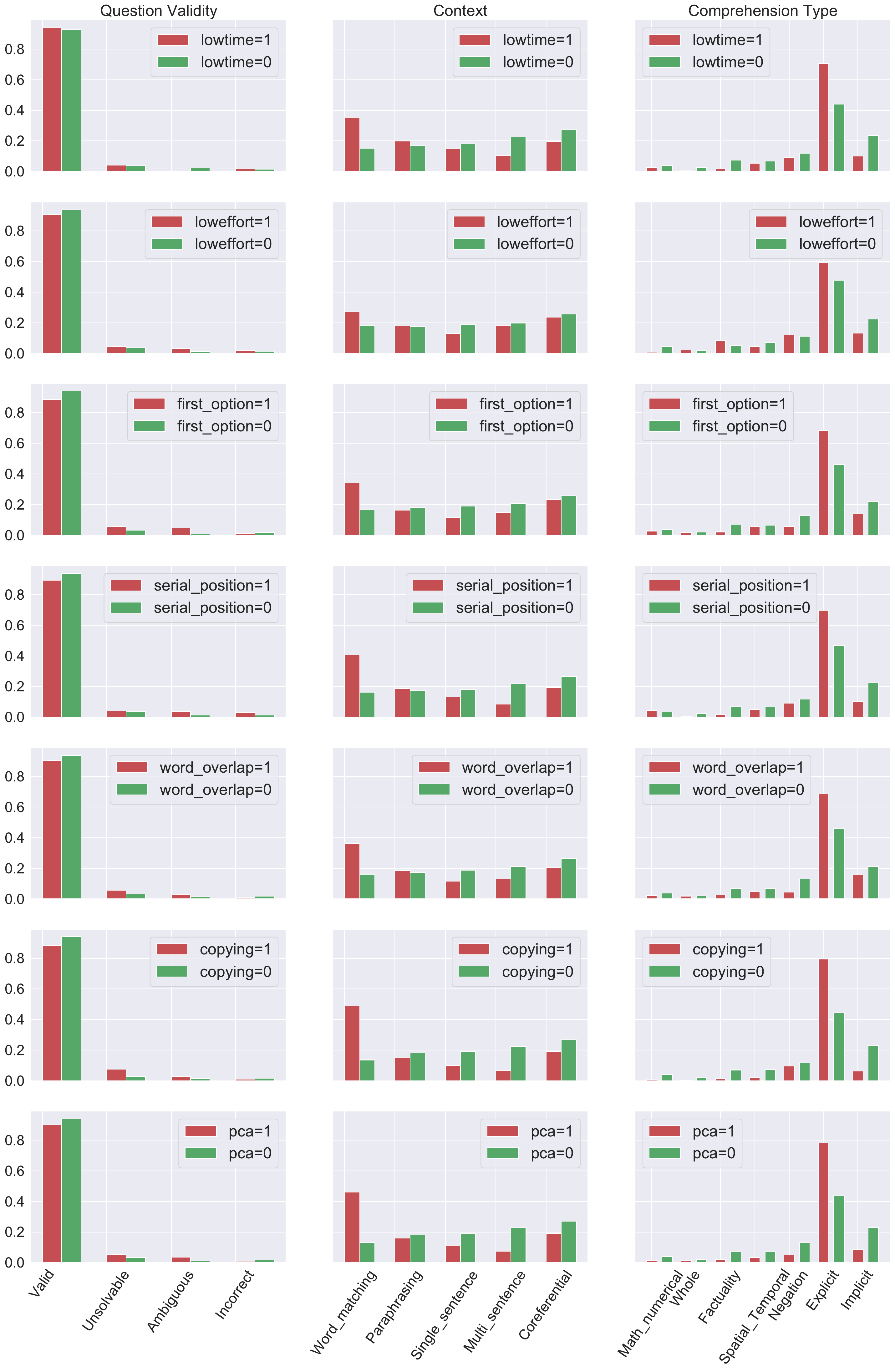}
    \caption{Breakdown of the question annotation results from section~\ref{sec:qual} when percentile \textit{k}=25.}
    \label{fig:ques_annot_full}
\end{figure*}

We describe the annotation scheme used to label examples for the analysis in section~\ref{sec:qual}. In addition, we show a breakdown of those results across all heuristic features in Figure~\ref{fig:ques_annot_full}.

\paragraph{Validity.} We annotate whether examples are answerable or not using the following labels:
\begin{enumerate}
  \item Unsolvable: It is not possible to answer the question given the context in the passage and question, or the question is underspecified or incoherent.
  \item Incorrect: The answer is marked incorrectly.
  \item Ambiguous: The question does not have a unique correct answer.
  \item Valid: The question can be reasonably answered from the passage.
\end{enumerate}

\paragraph{Context.} To understand how much context from the passage is needed to answer the question, we label questions using the following labels:
\begin{enumerate}
    \item Word matching: The question matches a span in the passage, and the answer is easily extractable by matching spans.
    \item Paraphrasing: The question paraphrases information in exactly one sentence in the passage, and the answer can be retrieved from it.
    \item Single-sentence reasoning: The question can be answered by exactly one sentence in the passage, but requires a conceptual overlap, or performing some other form of inference.
    \item Multi-sentence reasoning: The question can only be answered by synthesizing information from multiple sentences in the passage. This excludes just performing coreference.
    \item Coreferential Reasoning: The question requires performing coreference.
\end{enumerate}

\paragraph{Comprehension Type.} To determine what kinds of comprehension are required to answer the question, we label examples with the following labels:
\begin{enumerate}
    \item Math/numerical: Questions that require mathematical or numerical reasoning.
    \item Whole: Questions that require a complete understanding of the passage or ask about the author's opinion on the passage.
    \item Factuality: Questions asking about truthfulness of the statements presented in the question or the options (e.g., questions of the form "which of the following is true / false?", or "is it true/false that ..")
    \item Spatial/temporal: Requires understanding of location and temporal order of events.
    \item Explicit: Asks about information (facts, events or entities) stated in the passage explicitly or in a paraphrased manner. These shouldn't require much of a concept jump.
    \item Implicit: Asks about information not directly stated, but which can be inferred through commonsense, causality, numerical or other types of inference.
    \item Negation: Questions which are phrased in the form of a negation (for e.g. using keywords like "not" and "without").
\end{enumerate}

\end{document}